\documentclass{article}

\usepackage{microtype}
\usepackage{graphicx}
\usepackage{subfigure}
\usepackage{booktabs} 

\usepackage{mathtools}

\usepackage{hyperref}

\usepackage[accepted]{icml2021}

\icmltitlerunning{NEVAE}

\begin{document}

\twocolumn[
\icmltitle{Neighbor Embedding Variational Autoencoder}



\icmlsetsymbol{equal}{*}

\begin{icmlauthorlist}
\icmlauthor{Renfei Tu}{equal,to}
\icmlauthor{Yang Liu}{equal,to}
\icmlauthor{Yongzeng Xue}{to}
\icmlauthor{Cheng Wang}{to}
\icmlauthor{Maozu Guo}{goo}
\end{icmlauthorlist}

\icmlaffiliation{to}{Department of Computing, Harbin Institute of Technology, Harbin, China}
\icmlaffiliation{goo}{Beijing University of Civil Engineering and Architecture, Beijing, China}

\icmlcorrespondingauthor{Yang Liu}{liuyang@hit.edu.cn}
\icmlcorrespondingauthor{Renfei Tu}{renfei.tu@hit.edu.cn}

\icmlkeywords{Machine Learning, ICML}

\vskip 0.3in
]



\printAffiliationsAndNotice{\icmlEqualContribution} 

\begin{abstract}
Being one of the most popular generative framework, variational autoencoders(VAE) are known to suffer from a phenomenon termed posterior collapse, i.e. the latent variational distributions collapse to the prior, especially when a strong decoder network is used.
In this work, we analyze the latent representation of collapsed VAEs, and proposed a novel model, neighbor embedding VAE(NE-VAE), which explicitly constraints the encoder to encode inputs close in the input space to be close in the latent space.
We observed that for VAE variants that report similar ELBO, KL divergence or even mutual information scores may still behave quite differently in the latent organization. 
In our experiments, NE-VAE can produce qualitatively different latent representations with majority of the latent dimensions remained active, which may benefit downstream latent space optimization tasks.
NE-VAE can prevent posterior collapse to a much greater extent than it's predecessors, and can be easily plugged into any autoencoder framework, without introducing addition model components and complex training routines.
\end{abstract}

\section{Introduction}
\label{introduction}
Variational Autoencoders (VAE) \cite{kingma2013auto,rezende2014stochastic} and Generative Adversarial Networks (GAN) \cite{goodfellow2014generative} are arguably the most popular generative modeling framework that have seen wide applications on many tasks. Although it's commonly believed that VAEs are less effective in generating realistic samples than GANs \cite{dai2019diagnosing}, VAEs still remain very attractive for many researchers because it's architecture are naturally suited for a class of problem called Latent Space Optimization (LSO) \cite{tripp2020sample}, in which optimizations are performed in the low-dimensional, continuous latent manifold learned by a deep generative model, when optimizing an expensive black-box objective function over a complex, high-dimensional, and structured input space is unfeasible. Such method were proven promising in automatic chemical design \cite{gomez2018automatic,kusner2017grammar}, natural scene understanding \cite{lu2018structured}, molecular graph generation \cite{jin2018junction}, etc.

In the case of LSO, generating high-quality samples \(\hat{x}\) from the model \(p(x)\) is not the only concern, because we also expect the deep generative model to learn a 'well behaved' low-dimension representation \(p(z|x)\) of the high-dimension input data. Unfortunately, as pointed out in \cite{alemi2018fixing}, the loss function of VAE only depends on \(p(x|\theta)\), and not on \(p(x,z|\theta)\), thus they do not measure the quality of the low dimensional representation at all. Empirically, if we have a powerful stochastic decoder \(p(x|z,\theta)\), such as an RNN or PixelCNN, a VAE can easily be trained to ignore \(z\) and still obtain high marginal likelihood \(p(x|\theta)\), which phenomenon later termed \textit{posterior collapse} \cite{lucas2019understanding}. It was observed in \cite{bowman2015generating} for sentence generation and in \cite{chen2016variational} for image generation. Thus obtaining a good marginal likelihood is not enough for good representation learning.

Since the discovering of the posterior collapse phenomenon, a lot of research was devoted to the analysis and treatment of this effect which may hinder VAEs' application on latent representation learning.\cite{alemi2018fixing} Because the numerical symptom of posterior collapse being the vanishing of KL-divergence term, which directly encourages the variational distribution \(p(z|x)\) to match the prior \(p(z)\), earlier lines of research on this phenomenon naturally put the focus on limiting the effect of KL term in the ELBO objective. Following this line of heuristic, the most common approach for alleviating posterior collapse is to anneal a weight on the KL term during training from 0 to 1 \cite{bowman2015generating,sonderby2016ladder}. However, it requires a hand-crafted annealing schedule that's problem and model dependent. It's also shown that posterior will usually collapse again once the annealing is done\cite{lucas2019understanding}.

More recent researches on posterior collapse from multiple perspectives gave alternative interpretations and solutions for this phenomenon. \cite{higgins2016beta}, proposed the \(\beta\)-VAE for unsupervised learning of disentangled representations. Based on which, \cite{alemi2018fixing} furthered an analysis incorporating rate-distortion theory to change \(\beta\) for actively trade-off between reconstruction quality and sampling accuracy. \cite{theis2015note} proposed to inject uniform noise into the pixels or reduce the bit-depth of the data for image autoencoders. \cite{rezende2018taming} introduced extra constraints to adaptively balance between reconstruction and KL term rather than using a fixed \(\beta\). \cite{rainforth2018tighter} analyzed the training dynamics of importance weighted autoencoders (IWAE) and reported that using higher particle count for weighted sampling can make the ELBO tighter and the decoder more powerful, at the cost of a lower signal-to-noise ratio for the gradient updates for the encoder network. They proposed to use separated training objectives for the encoder and decoder networks. \cite{he2019lagging} argued that training trajectory of the encoder network lagging behind the decoder is partially responsible for causing posterior collapse, and proposed to aggressively train the encoder network for more iterations to maximize mutual information between the input and the latent. \cite{razavi2019deltavae} proposed to use a lower bound constraint on the KL term. \cite{lucas2019understanding} provided theoretical analysis on the correspondence between linear VAEs and Probabilistic PCA, and empirical evidence for the presence of local minima causing posterior collapse in non-linear VAEs. \cite{dai2020reassessing} gave alternative theoretical analysis on why posterior collapse is not only in consequence of the KL term, but also bad local minima inherent to the loss surface of deep autoencoder networks.

In this work, we:
\begin{itemize}
    \item Proposed Neighbor Embedding Variational Autoencoder (NE-VAE), by penalizing the distance between a latent code and its re-encoding, which effectively distributes activity evenly between latent dimensions, and prevents posterior collapse. It's easy to plug into any existing VAE framework for downstream tasks.
    \item We intuitively explained NE-VAE's ability of preventing posterior collapse, and provided objective numerical evidence for NE-VAE's effectiveness in preventing posterior collapse. We performed latent traverse analysis to demonstrate the qualitative feature of active and inactive latent dimensions.
    \item A toy latent space optimization task was used to demonstrate the potential benefits of NE-VAE for LSO tasks. Optimization on the latent representation of NE-VAE is less susceptible to local minima, and converges faster.
    \item We provided empirical evidence that the distribution of activity in latent dimensions can be affected by constraints and even data configurations, which may not be reflected by available simple numerical metrics.
\end{itemize}

\section{Background}

\subsection{Variational Autoencoder}

VAE is a popular deep latent variable model. Its generative model was defined as a marginal distribution of the observations \(x\in \mathcal{X} \):
\begin{equation}
    \label{data likelihood}
    p_\theta(x)=\int p_\theta(x|z)p(z)dz    
\end{equation}

where the decoder \(p_\theta(x|z)\) is parameterized with deep neural network, and \(p(z)\) is the prior distribution of the latent variable, usually just unit Gaussian. Because this marginal likelihood is usually intractable, a variational distribution \(q_\phi(z|x)\) is introduced as the variational surrogate for the true posterior \(p(z|x)\). This enabled us to optimize a tractable variational lower bound(ELBO) of the data likelihood \(\log p_\theta(x)\):
\begin{equation}
    \label{elbo}
    \begin{split}
        \mathcal{L}(x;\theta,\phi)=&\\
        E_{z\sim q_\phi(z|x)}&[\log p_\theta(x|z)]-D_{KL}[q_\phi(z|x)||p(z)]        
    \end{split}
\end{equation}

where encoder \(q_\phi(z|x)\) and decoder \(p_\theta(x|z)\) are both parameterized with deep neural network with \(\phi\) and \(\theta\) as their respective parameters. Optimize this lower bound will push the variational distribution \(q_\phi(z|x)\) to approximate the true posterior \(p_\theta(z|x)\). This lower bound is composed of two competing terms. Intuitively, the first half is the \textit{reconstruction loss} that push the model to encode the necessary information to reconstruct the input data as good as possible, and the second part is regularizing the variational distribution \(q_\phi(z|x)\) towards the prior \(p(z)\). In the scope of this paper, we assume \(p(z)=\mathcal{N}(\textbf{0}, \textbf{I})\) unless otherwise specified.

\subsection{Posterior Collapse}

There are now abundant researches that give various explanations and solutions to the phenomenon of posterior collapse as mentioned in the introduction. At this stage, it maybe untenable or impractical trying to isolate an singular cause and fix for posterior collapse that fits all applications.

It maybe better to connect the phenomenon and application, and use symptomatic treatments that fits the goal of the overall task at hand. To address the ambiguity of the definition, \cite{dai2020reassessing} gave an useful taxonomy of different scenarios that were all referred to as posterior collapse. And it's argued that selective posterior collapse may even be desirable for some tasks \cite{dai2019diagnosing}. Indeed, for generic generative tasks, if we can get good reconstruction and samples, posterior collapse shouldn't be a concern. But if we want to build VAE as a predecessor for downstream latent space optimization tasks that expect the model to learn a latent manifold with desirable properties for LSO, we have a problem. 

In this paper, we describe posterior collapse as the phenomenon when some or all latent dimensions of \(z\) are collapsed to the prior for all inputs, i.e.:

\begin{equation}
    \label{posterior collapse}
    \forall x, \exists j:\quad q_\phi(z_j|x)\approx p(z_j)=\mathcal{N}(0,1)
\end{equation}

In the scenario of complete posterior collapse, the model can't pass information from the input to the decoder, thus no meaningful latent representation can be learned at all. In a more common setting, a large part of the latent dimension will be collapsed. Simply adding more dimensions to the latent space may not affect how many dimensions will be preserved from collapsing after the training\cite{dai2019diagnosing}, which behavior will be confirmed in our experiments.  This is undesirable if we want VAE to learn a meaningful low dimensional representation of the input in a controlled fashion. As we'll see in the experiments, a latent representation with too many collapsed dimensions may cause latent space optimization tasks to get stuck in local minima, and slow down the optimization process.

\section{Related Works}


\subsection{KL-oriented Methods}

As mentioned in the introduction, majority of the works on the posterior collapse phenomenon are focused on the KL term, naturally as a result of the fact that posterior collapses when KL term vanishes. 

But as per the analysis by \cite{alemi2018fixing} from the information-theoretic perspective, when the network is powerful enough, there may exist plenty of different solutions that can't be distinguished by the ELBO objective alone. 
At the extreme, the ELBO objective may not distinguish between models that make almost no use of the latent variable, versus models that make large use of the latent variable and learn useful representations for reconstruction. 
Based on this analysis, and the observation that ELBO objective don't measure the quality of the learned latent representation\cite{alemi2018fixing}, we hypothesize that for similar ELBO scores there exists different model parameters that correspond to qualitative different latent representations.
In this work, we advance this reasoning to finer granularity, and provide empirical evidence that qualitatively different latent representations not only exist for ELBO, but also arises even when we further fix both reconstruction loss and KL term. i.e. we can have qualitatively very different latent organization that reaches similar reconstruction and KL losses. 
Thus, we'll need extra constraints on the latent organization to reach optimal result that suits specific task needs.

    \subsubsection{KL Annealing}
    KL annealing gives the KL term a smaller weight during the course of training and gradually fall back to vanilla VAE. 
    Introduced by \cite{bowman2015generating,sonderby2016ladder}, KL annealing may partially alleviate posterior collapse when combined with other techniques. 
    But the application of KL annealing requires hand-crafted annealing schedule, and it's reported that posterior may collapse again after the annealing phase, thus it's insufficient for preventing posterior collapse when used alone\cite{lucas2019understanding}.
    It's observed that with different annealing schedules, the model will converge to different ELBO values \cite{lucas2019understanding}, which is consistent to other analysis that attribute posterior collapse partially to bad local minima \cite{dai2020reassessing}. 

    \subsubsection{\(\beta\)-VAE} 
    Proposed by \cite{higgins2016beta}, \(\beta\)-VAE scales the KL term with an extra parameter \(\beta\):
    \begin{equation}
        \label{beta-loss}
        \begin{split}
        \mathcal{L}_\beta(x;\theta,\phi)=&\\
        E_{z\sim q_\phi(z|x)}&[\log p_\theta(x|z)]-\beta D_{KL}[q_\phi(z|x)||p(z)]
        \end{split}
    \end{equation}
    
    Originally designed for learning disentangled representations, it was utilized by \cite{alemi2018fixing} for an analysis to use \(\beta\) to trade off between reconstruction quality and sampling accuracy, where a smaller \(\beta\) can prevent posterior collapse but at the cost of higher KL penalty and deteriorated sampling quality. According to \cite{tolstikhin2017wasserstein}, the deterioration of sampling quality may be a result of aggregated posterior deviating too much from the prior.

\subsection{Aggressive Training} 

Different from common KL-oriented solutions, \cite{he2019lagging} try to tackle posterior collapse from the perspective of training dynamics. They provide empirical evidence that during the training process, the true model posterior is a moving target, while the training of the encoder network can't catch up with the decoder network. As a result, the encoder is encouraged to ignore the latent code and collapse to the prior. They distinguished the collapse of variational posterior from the collapse of true model posterior, and proposed to aggressively train the encoder network for more iterations to alleviate this lagging effect. This finding is both consistent with \cite{rainforth2018tighter} that a powerful decoder network can cause the gradient update of the encoder network to deteriorate, and with the works suggesting the abundance of local minima present in VAE's parameter space\cite{dai2019diagnosing,dai2020reassessing}, because aggressive training schedule don't change the objective while reaching different minima than vanilla VAEs.

\section{Method}

\subsection{Intuition}

Although VAE is generally considered to be performing dimension reduction and information compression automatically, but as already discussed above, the ELBO objective can't guarantee the learned latent representation to be organized in certain desirable way among other local minima. Indeed, uncontrolled posterior collapse can easily happen for vanilla VAE, preventing the latent dimensions to be fully used, or used equally. This work aims to resolve this issue by forcing VAE to produce certain desirable latent representation.

Inspired by the successful dimension reduction and high-dimensional data visualization technique, Stochastic Neighbor Embedding\cite{hinton2002sne}, we propose to enforce additional constraint on the learned representation, i.e. the learned low-dimensional latent manifold should maintain the topological features of the input manifold as much as possible. In this case, we would love the the latent codes \(z_1, z_2 \in \mathcal{Z}\) of two input sample \(x_1, x_2 \in \mathcal{X}\) to be close to each other in \(\mathcal{Z}\) if \(x_1, x_2\) are close to each other in \(\mathcal{X}\).

If this can be achieved, we can expect the learned latent space to be much more regular than other ill-formed solutions, and potentially reduce bad local minimal solutions for downstream tasks. Less obviously, this type of constraint may also prevent posterior collapse from happening.

When posterior collapse happens, latent code of the collapsed dimensions will be almost equivalently sampled from the latent prior. In which case, no information of the inputs will be passed to the latent code, i.e. there's a good chance that \(z_1, z_2\), the latent code of two input vector \(x_1, x_2\) that's similar, or close to each other in \(\mathcal{X}\), will be distant in \(\mathcal{Z}\) because they're scattered according to a unit Gaussian. This distance will be punished by our "neighbor preserving" constraints.

However, we can't straightforwardly implement this constraint by pushing the latent code of existing input vectors to conform their configuration in the input space \(\mathcal{X}\), because that would require us to 1) use an additional metric to measure the similarity between input vectors, which is not exactly obvious, and 2) remember the locations the sample's neighbors in \(\mathcal{X}\) were encoded to, when we're processing every sample. And since the parameter of the network is being updated during the course of training, we may have to pass every neighbor sample through the encoder in every iteration, which would be computationally unfeasible.

Fortunately, since the decoder network of VAE will naturally be pushed to generate samples that's similar to the input that was encoded, we can use those generated samples as surrogates for those neighbor samples. When we're processing input \(x_i\in \mathcal{X}\) that was encoded to the latent code \(z_i\), the decoder will produce reconstruction \(\hat{x_i}\) which will be taken as neighbors of \(x_i\) in \(\mathcal{X}\). Then we can put the 'artificial neighbor' into the encoder again and \textit{re-encode} it into latent code \(\hat{z_i}\). Since the reconstruction error term of VAE objective already pushes \(x_i\) and \(\hat{x_i}\) to be close, our additional constraint only need to further push two respective latent code, \(z_i\) and \(\hat{z_i}\) to be close to each other.

\subsection{Enforcing Neighbor Embedding Constraint by Re-encode Reconstruction}

In conventional VAE practice, we find the parameter for the encoder and decoder networks by optimizing over ELBO (Eq. \ref{elbo}):

\begin{equation}
    \label{opt-elbo}
    {\arg\max}_{\theta,\phi} \mathcal{L}(x;\theta,\phi)
\end{equation}

To put the above explained intuition into implementation, we add additional \textit{re-encoding loss} to the ELBO objective.

For a given data sample \(x\), we can get its corresponding stochastic latent code \(z\sim q_\phi(z|x)\) from the encoder, and its reconstruction \(\hat{x}\sim p_\theta(x|z)\) from the decoder. We put this reconstruction into the encoder again to get the re-encoded latent code \(\hat{z}\sim q_\phi(z|\hat{x})\). Then we can use different method to penalize the distance between \(z\) and \(\hat{z}\):

\begin{equation}
    \label{ne-loss}
    {\arg\max}_{\theta,\phi} \mathcal{L}(x;\theta,\phi) - \mathcal{L}_{NE}
\end{equation}

where \(\mathcal{L}_{NE}\) can be a fixed squared error loss that's symmetric to the conventionally used reconstruction loss in ELBO \cite{dai2019diagnosing}:

\begin{equation}
    \label{ne-se}
    \mathcal{L}_{NE}^{se}(z,\hat{z}) = \sum_{i=1}^{n_z}(z_i-\hat{z}_i)^2
\end{equation}

where \(n_z\) is the number of latent dimensions. Or alternatively, we can use an adjustable log probability loss:

\begin{equation}
    \label{ne-lp}
    \mathcal{L}_{NE}^{lp}(z,\hat{\mu}, \hat{\sigma})=\sum_{i=1}^{n_z}(L_{NE}^{lp}(z, \hat{\mu}, \hat{\sigma}))_i
\end{equation}

where

\begin{equation}
    \label{lp-capping}
    (L_{NE}^{lp}(z, \hat{\mu}, \hat{\sigma}))_i=
    \begin{cases}
        0, \text{  \quad   if } c>-\log p(z_i;\hat{\mu}_i, \hat{\sigma}_i) \\
        -\log p(z_i;\hat{\mu}_i, \hat{\sigma}_i), \text{     otherwise} 
    \end{cases}
\end{equation}

in which \(p(\cdot ;\hat{\mu}_i, \hat{\sigma}_i)\) is the pdf of \(\mathcal{N}(\hat{\mu}_i,\hat{\sigma}_i^2)\), and \(c\) is the capping hyper parameter to prevent unbounded optimization. This loss follows a similar intuition that a latent code should be assigned a high probability under the re-encoded variational distribution.

Effectively, we're dropping the constraint and fall back to vanilla VAE if the respective dimension of \(z\)'s negative log likelihood is already smaller than \(c\) under \(\mathcal{N}(\hat{\mu}_i,\hat{\sigma}_i^2)\). Thus we can adjust \(c\) to control how packed the 'neighborhood' should be in the latent space. If we use a \(c\) that's too big, the constraint will always be dropped thus the whole model falls back to vanilla VAE. If we use a  \(c\) that's too small, the model will learn to encode samples to variational distributions with very small variance, thus break the continuity of the latent space and result in poor sampling quality. The case of uncapped log probability loss (which is equivalent to \(c=-\infty\)) will drive the encoder to produce variational distribution of ever smaller variance, thus reaching arbitrary small negative log likelihood re-encoding loss \(\mathcal{L}_{NE}^{lp}\).

\section{Experiments}

\subsection{Setup}

For all experiments, we use a unit Gaussian prior \(\mathcal{N}(\textbf{0},\textbf{I})\), and the encoder network produces mean and log variance for the diagonal Gaussian variational distribution \(q_\phi(z|x)\). We report negative ELBO, \(D_{KL}[q_\phi(z|x)||p(z)]\) (KL), mutual information \(I_q\) (MI)\cite{hoffman2016elbo}:
\begin{equation}
    \label{mutual-information}
    I_q=\mathrm{E}_{x\sim p_d(x)}[D_{KL}(q_\phi(z|x)||p(z))]-D_{KL}(q_\phi(z)||p(z))
\end{equation}
where \(p_d(x)\) is the empirical distribution. The aggregated posterior, \(q_\phi(z)=\mathrm{E}_{x\sim p_d(x)}[q_\phi(z|x)]\) and how far it divergent from the prior \(D_{KL}(q_\phi(z)||p(z))\) can be approximated with Monte Carlo estimate. We also report the number of active units (AU) \cite{burda2016importance} in latent representation. The activity of a latent dimension \(z\) is measured as \(A_z=\mathrm{Cov}_x(\mathrm{E}_{z\sim q(z|x)}[z])\) and the dimension is regarded as active when \(A_z>0.01\). For the purpose of different discussions, we'll provide additional measurements including per-dimension activity \(A_z\), re-encoding loss (\(\mathcal{L}_{NE}^{se}\) or \(\mathcal{L}_{NE}^{lp}\)). All models are trained with a fixed schedule of 100 epochs with same random seed. KL annealing are all scheduled to anneal from 0.1 to 1 in the first 10 epochs unless otherwise specified.

To fully characterize the behavior or latent space produced by different models, we'll apply latent traverse from multiple perspectives, as detailed below.

For baselines, we report results form vanilla VAE\cite{kingma2013auto}, \(\beta\)-VAE\cite{higgins2016beta} and aggressive training\cite{he2019lagging}. Our network architecture follows the configuration of \cite{he2019lagging} for image data, that the encoder is implemented with ResNet\cite{he2016resnet} and decoder with a 13-layer Gated PixelCNN\cite{oord2016pixelcnn}. The annealing schedule for \(\beta\)-VAEs are the same with other models other than that the KL weight stops annealing once it arrives at the respective \(\beta\) value.

We evaluate the models with Omniglot data set\cite{lake2015omniglot} and conducted comparative experiments against MNIST\cite{lecun1998mnist}, which is qualitatively similar to Omniglot, but contains much more samples per class, and is composed of only 10 classes. We use 32-dimensional latent code \(z\) and optimize the objective with Adam optimizer\cite{kingma2015adam}

\subsection{Numerical Results}

In table-\ref{table-main} we show the main numerical results on Omniglot with fixed latent dimension count \(n_z\)=32, our method can achieve better posterior collapse preventing effect than listed baselines. Specifically, proposed squared error version of NE-VAE reached higher MI and kept most of the latent dimensions active. Comparing to \(\beta\)-VAE with \(\beta\)=0.9, our proposed model achieved comparable MI and AU with significantly less KL penalty. From equation-\ref{mutual-information}, we can intuitively understand the gap between mutual information and KL loss as the divergence between aggregated posterior \(q_\phi(z)\) and prior \(p(z)\). As discussed in \cite{tolstikhin2017wasserstein, he2019lagging}, this gap is undesirable because when aggregated posterior divergent too much away from the prior, the model will likely have bad sampling quality. \cite{alemi2018fixing} analyzed ELBO from the perspective of rate-distortion theory and treated the KL term as the bandwidth of the channel, which is a upper bound of the actually transmitted information. Simply giving the KL term a smaller weight to forcefully enlarge the bandwidth, may not ensure the bandwidth to be used meaningfully by the autoencoder to encode useful information, hence the big gap between MI and KL for \(\beta\)-VAE. Log probability versions of NE-VAE can achieve similar effect for AU but have lower MI, which behaves similar to \(\beta\)-VAE.

In table-\ref{table-nz} we report AU count for VAE, aggressive training method and NE-VAE se with varying \(n_z\) of 8, 16 and 64. For vanilla VAE and aggressive training method, the reported AU count are largely independent of the latent dimensional count hyper parameter, while NE-VAE se can keep a large proportion of the latent dimensions active.

\begin{table}[t]
    \caption{Results on Omniglot. NE-VAE se achieved significant higher AU count than vanilla and aggressively trained VAE, indicating better posterior collapse preventing effects. Meanwhile, NE-VAE se don't suffer from the big gap between MI and KL as \(\beta\)-VAEs.}
    \label{table-main}
    \vskip 0.15in
    \begin{center}
    \begin{small}
    \begin{sc}
    \begin{tabular}{lcccc}
    \toprule
    Method & -ELBO & KL & MI & AU \\
    \midrule
    VAE no anneal                &88.79 &1.10 &1.08 &2 \\
    VAE                          &88.91 &2.12 &1.83 &6 \\
    Aggressive                   &88.71 &1.80 &1.64 &5 \\
    Aggr. no anneal              &88.84 &2.38 &2.28 &6 \\
    \(\beta\)-VAE(\(\beta=\)0.9) &89.42 &5.48 &3.32 &27 \\
    \(\beta\)-VAE(\(\beta=\)0.8) &90.92 &16.25 &3.86 &32 \\
    \(\beta\)-VAE(\(\beta=\)0.7) &93.65 &27.56 &3.87 &32 \\
    \(\beta\)-VAE(\(\beta=\)0.6) &96.77 &37.12 &3.87 &32 \\
    \(\beta\)-VAE(\(\beta=\)0.5) &100.62&46.37 &3.87 &32 \\
    NE-VAE se                    &89.57 &\textbf{2.55} &\textbf{2.39} &\textbf{26} \\
    NE-VAE lp(c=5.0)             &88.93 &2.01 &1.33 &6 \\
    NE-VAE lp(c=3.0)             &88.89 &1.75 &1.19 &6 \\
    NE-VAE lp(c=1.0)             &88.96 &1.69 &0.47 &6 \\
    NE-VAE lp(c=0.1)             &89.61 &3.40 &0.81 &28 \\
    NE-VAE lp(c=0.0)             &89.64 &4.02 &0.87 &22 \\
    NE-VAE lp(c=-0.1)            &89.89 &5.23 &1.21 &28 \\
    NE-VAE lp(c=-1.0)            &94.33 &19.28 &3.71 &32 \\
    NE-VAE lp(c=-3.0)            &122.03&60.03 &3.87 &32 \\
    NE-VAE lp(c=-5.0)            &148.33&87.50 &3.87 &32 \\
    \bottomrule
    \end{tabular}
    \end{sc}
    \end{small}
    \end{center}
    \vskip -0.1in
\end{table}

\begin{table}[t]
    \caption{Reported AU count with varying latent dimension size \(n_z\). Both vanilla VAE and aggressive training method give similar AU count regardless of the latent dimension hyper parameter. NE-VAE se can produce AU proportional to \(n_z\), keeping majority of the latent dimensions active. }
    \label{table-nz}
    \vskip 0.15in
    \begin{center}
    \begin{small}
    \begin{sc}
    \begin{tabular}{lccc}
    \toprule
    Method & AU(\(n_z\)=8) & AU(\(n_z\)=16) & AU(\(n_z\)=64) \\
    \midrule
    VAE         & 5 &7  &4  \\
    Aggressive  & 7 &7  &4  \\
    NE-VAE se   & 8 &12 &55 \\

    \bottomrule
    \end{tabular}
    \end{sc}
    \end{small}
    \end{center}
    \vskip -0.1in
\end{table}

\begin{figure}[ht]
    \begin{center}
        \centerline{\includegraphics[width=\columnwidth]{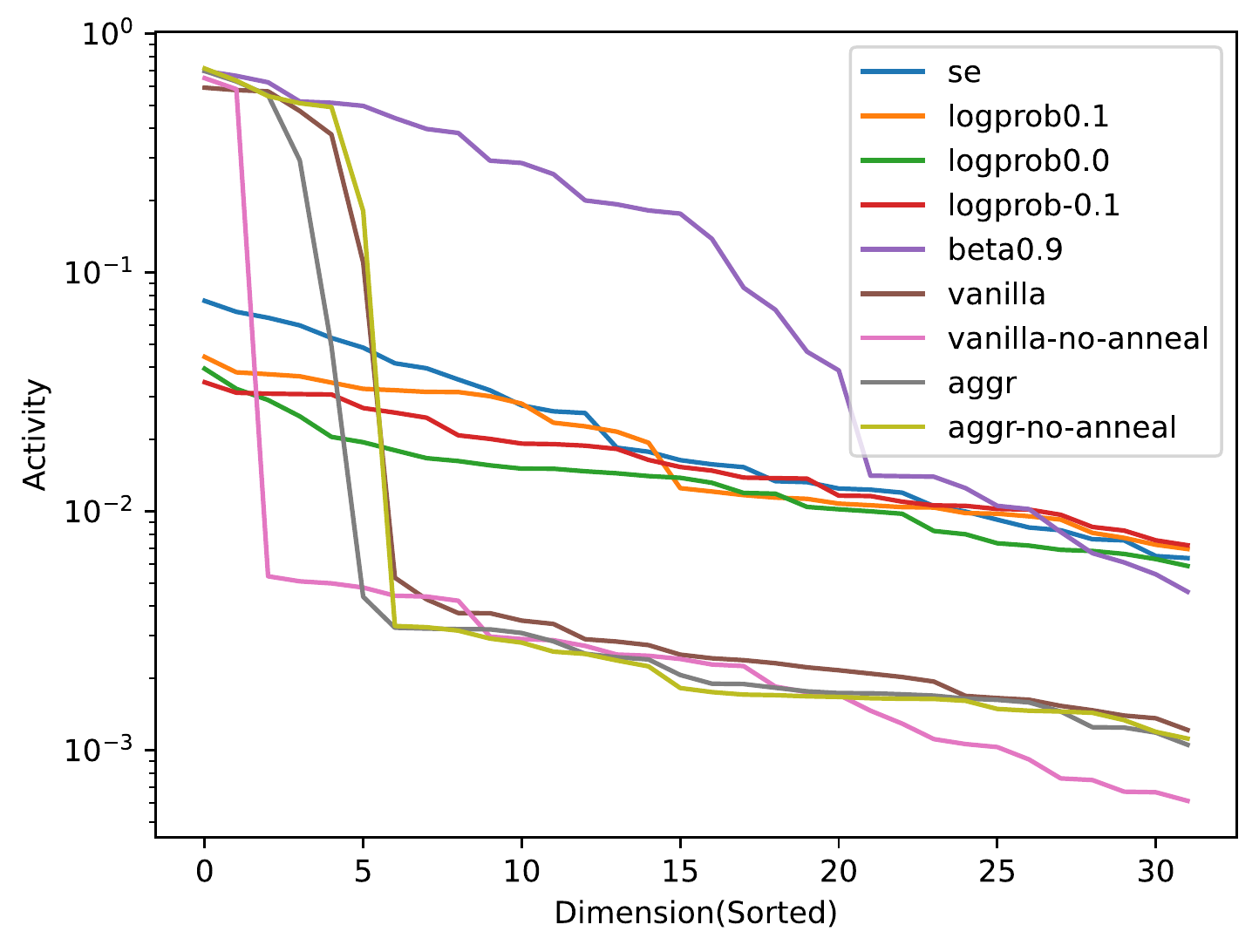}}
        \caption{Latent per-dimension activity \(A_z\)(in log scale). Dimensions are sorted w.r.t. \(A_z\) for better visualization. Our method distribute activity along different dimensions much more even than vanilla and aggressively trained VAEs, indicating NE-VAE can make use of most of the latent dimensions.}
        \label{per-dim-activity}
    \end{center}
\end{figure}


In addition to the reported AU count, we also report per-dimension latent activity \(A_z\) in fig-\ref{per-dim-activity}. We observed that both Vanilla VAE and aggressive training method have few very active latent dimensions while majority of the dimensions collapsed. NE-VAEs are capable of squashing the activity to be distributed more evenly in all latent dimensions, thus preventing majority of the latent dimensions from collapsing.

\begin{figure}[ht]
    \begin{center}
        \centerline{\includegraphics[width=\columnwidth]{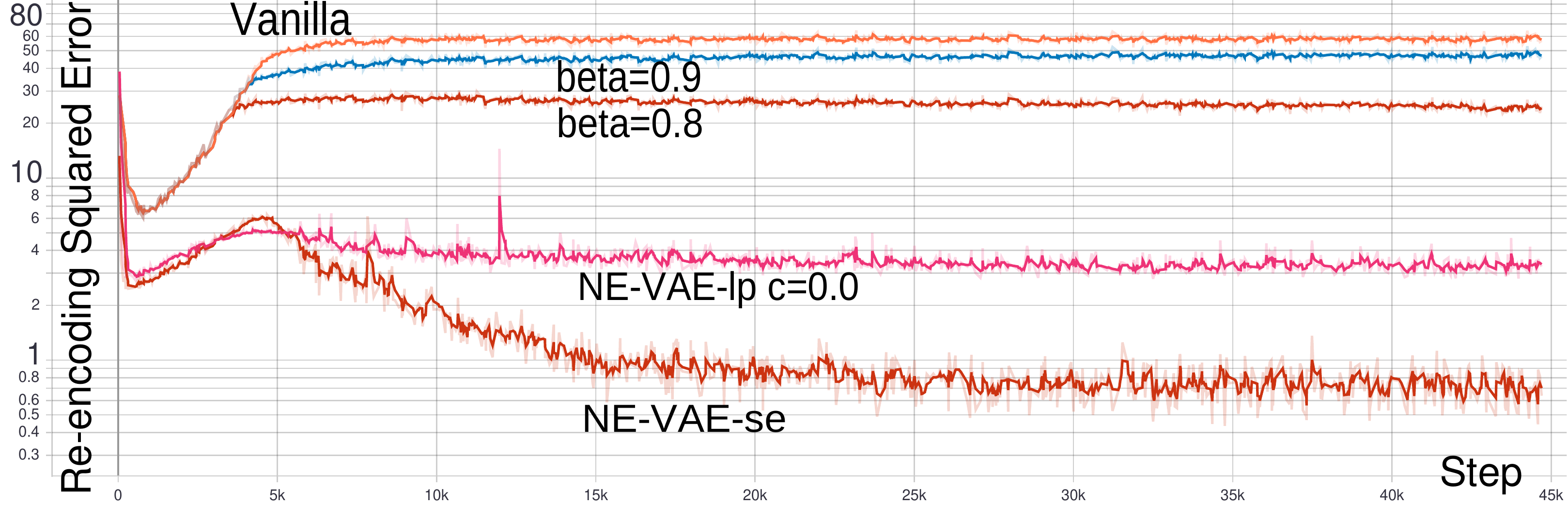}}
        \caption{Re-encoding squared error for VAE, \(\beta\)-VAE and NE-VAE. Without explicitly constraint on re-encoding error, reducing the weight on KL term can also reduce the reported re-encoding error, but the effect is much more attenuated than NE-VAE.}
        \label{se-training}
    \end{center}
\end{figure}

In fig-\ref{se-training}, we plot the reported re-encoding squared error for vanilla VAE, \(\beta\)-VAE and our methods. Since the annealing period of vanilla VAE can be seen as a \(\beta\)-VAE with gradually increasing \(\beta\), we observe that by reducing \(\beta\), that is to reduce the weight of the KL term, will also reduce the re-encoding error, but its effect is much more attenuated than explicitly constraining on the re-encoding loss. For vanilla VAE, once the annealing period is over, the re-encoding error reverts back to level even higher than untrained model. This result is consistent with a largely collapsed model since collapsed dimensions will have large re-encoding error. That is, for the collapsed dimensions, the encoding of the reconstruction \(\hat{x}\) will largely be detached from the encoding of \(x\), and the latent code for two similar inputs close in \(\mathcal{X}\), will be randomly scattered in \(\mathcal{Z}\), rather than close to each other. 

\subsection{Latent Traverse Analysis}
To visualize how change in the latent space affects the generated samples, we traverse the latent with multiple schemes and plot the respective generated samples.
\subsubsection{Single-dimension traverse}

\begin{figure}[ht]
    \begin{center}
        \centerline{\includegraphics[width=\columnwidth]{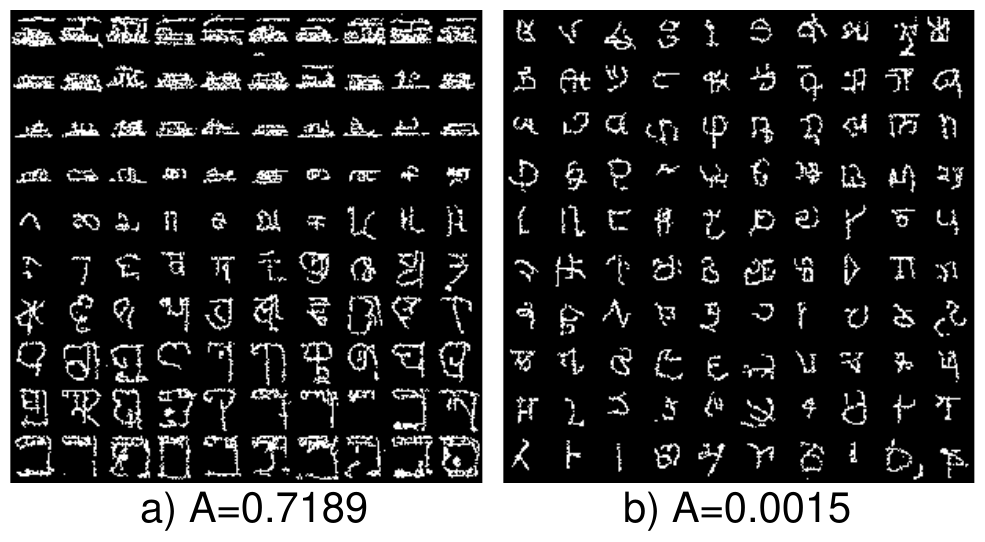}}
        \caption{Two dimensions of vanilla VAE (no annealing) with very different qualitative activity and their reported activity index.  (Best viewed enlarged on digital version) As shown in fig-\ref{per-dim-activity}, vanilla VAE only retains 2 very active dimensions like in a), while all other 30 dimensions are collapsed, reducing the traverse to close random behavior in those dimensions.}
        \label{lat-vae}
    \end{center}
\end{figure}

\begin{figure}[ht]
    \begin{center}
        \centerline{\includegraphics[width=\columnwidth]{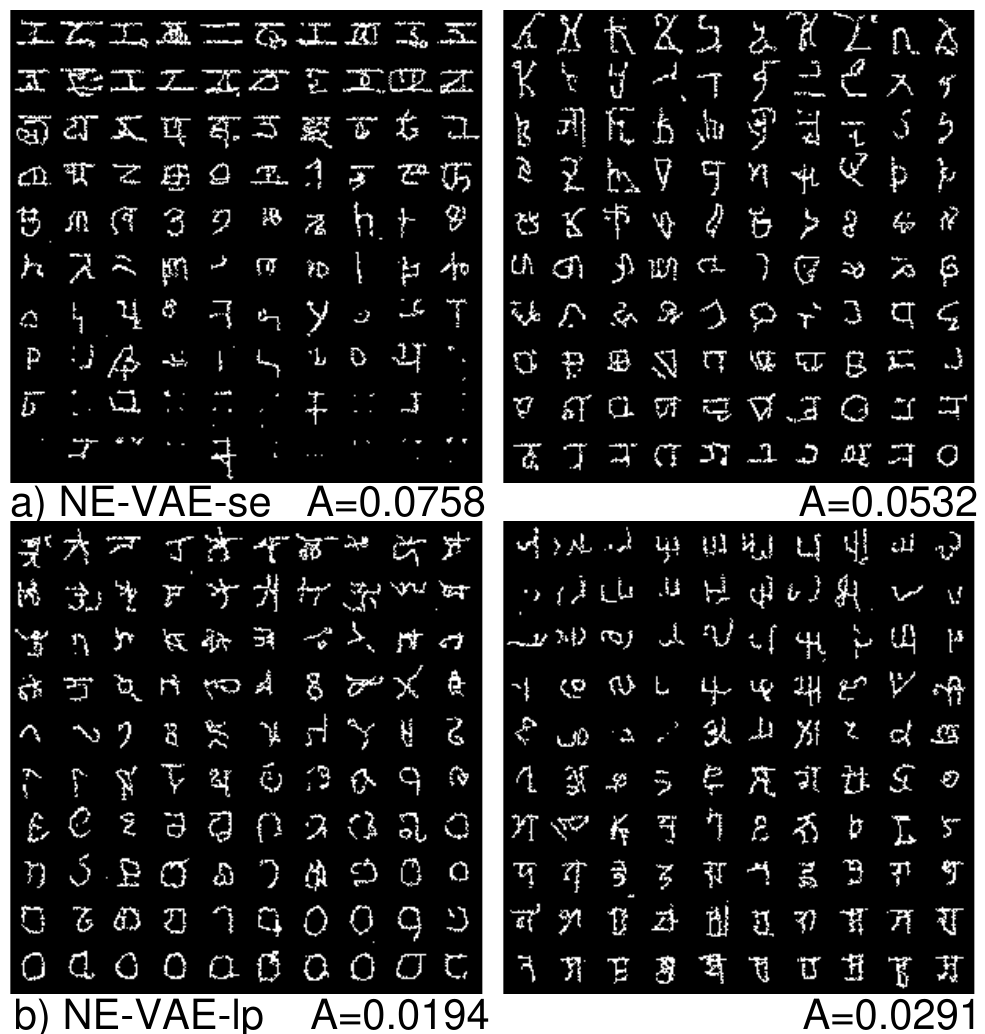}}
        \caption{Single-dimension interpolation for NE-VAE. Majority of the dimensions remained active.}
        \label{lat-nevae}
    \end{center}
\end{figure}

First, we interpolate on single dimension for 100 points between -10 and +10, while fixing all other dimensions to zero, to inspect the semantic activity of each latent dimension. In fig-\ref{lat-vae} we plot the traverse result of two latent dimensions of vanilla VAE(without KL annealing) that behave drastically different. \footnote{For a complete collection of plots from every latent dimension from all tested models please refer to the supplement materials.} Fig-\ref{lat-vae}a) shows a hyper active latent dimension which value have obvious correlation with the generated samples, while fig-\ref{lat-vae}b) shows a dimension which latent code is largely ignored by the decoder, or treated as random noises, thus produce interpolation traverse qualitatively similar to random samples. The contrast is easily distinguishable in fig-\ref{per-dim-activity} that we can count and see our trained vanilla VAE only have 2 very active dimensions, while others largely collapsed. In fig-\ref{lat-nevae} we plot two dimensions from two versions of NE-VAEs and report their respective activity index. By inspecting all traversed 32 dimensions of NE-VAE, we observe many more dimensions remained active than vanilla VAEs. Notably, NE-VAE-se produce latent space with majority of its dimensions semantically active, albeit some dimension's activity is more subtle than others. We also observe aggressive training method produce latent space behaves similar to vanilla VAE, which is consistent with the per-dimension activity reported in fig-\ref{per-dim-activity}.

\subsubsection{Random point interpolation}

\begin{figure}[ht]
    \begin{center}
        \centerline{\includegraphics[width=\columnwidth]{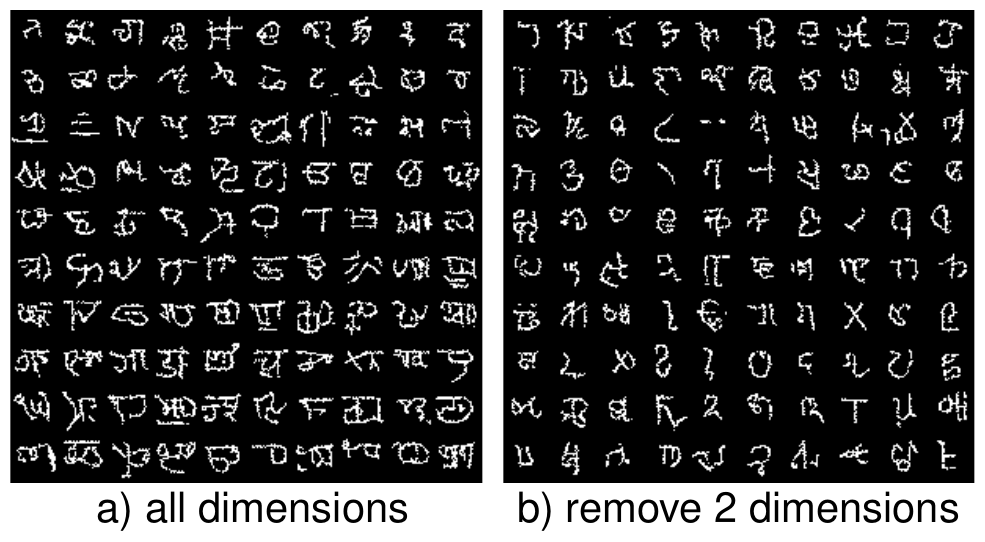}}
        \caption{Interpolate to random direction in the latent space of vanilla VAE. a) When all dimensions are used, the overall direction shows semantic activity. b) After fixing two previously identified active dimensions to zero, semantic activity is lost, and interpolation plots were reduced to random sampling.}
        \label{rm-dims}
    \end{center}
\end{figure}

To investigate if the two previously identified semantically active dimensions dominates the activity of vanilla VAEs, in this section, we randomly select a point in the latent space with fixed Euclidean distance 10 from the origin, and interpolate 100 points from the origin to the random target, which means will be smoothly changing all 32 latent dimensions rather than one. We show a typical result in fig-\ref{rm-dims}a)\footnote{For a larger collection of random trials, please refer to the supplement materials.}, we observed that almost all random trials consistently showed visible semantic activity. But if we fix the two active dimensions to zero out of all 32 dimensions and repeat the process, the result consistently shows an random sampling-like behavior without visible semantic activity, as shown in fig-\ref{rm-dims}b). This suggest that in a vanilla VAE with majority of the latent dimensions collapsed, the few not collapsed dimensions will dominate most of the semantic information, while collapsed dimensions are treated like random noises, reflecting it's near-zero bandwidth.

\subsubsection{Different behavior on MNIST}



We repeat the above traverse experiments on same models trained on MNIST, and observed that vanilla VAE (no annealing) have at least 5 obvious visually active latent dimensions, which is better than on Omniglot.

\subsection{Toy Latent Space Optimization Task}

To demonstrate if latent representation produced by different models will affect latent space optimization tasks, we tested NE-VAE se, vanilla VAE, aggressive training method and \(\beta\)-VAE with a toy LSO task.\footnote{Detailed statistics are available in the supplement materials.} 50 images \(x_i\) were sampled from the Omniglot data set. For every model, we freeze the parameter of their decoder, and treat the latent code \(z_i\) as the parameter for optimization. \(z_i\) were initialized with random value and \(\mu_i\) produced by the encoder in different trials. We use squared error between the ground truth \(x_i\) and the decode result \(\hat{x_i}\) as the loss, and an Adam optimizer was used to optimize \(z\) to make the decode result match the ground truth image. We stop the optimization when the change in loss is smaller than a threshold.

We observed that NE-VAE se are less sensitive to different initialization regardless of stopping threshold, the loss will consistently converge to much closer values than other listed models, especially when the stopping threshold is high(0.01-0.005). We also measure the squared Euclidean distance between codes converged to after different initialization. As we reduce the stopping threshold, the codes optimized in the latent space of NE-VAE se consistently getting closer to each other. The distance between codes being optimized actually diverges further away from each other for both vanilla VAE and aggressive training method, indicating a sensitivity to initialization. In the case of \(\beta\)-VAE(\(\beta=0.9\)), the distance undergoes a process of rise and fall as we reduce the stopping threshold, but the end distance remains larger than NE-VAE se.

In addition to the insensitivity to different initialization and more consistent convergence, optimization in the representation from NE-VAE se also converges the fastest when we set the stopping threshold to 0 ($\approx$16200 iterations, as opposed to $\approx$26900 iterations for vanilla VAE, $\approx$35900 iterations for aggressively trained VAE and $\approx$21600 iterations for \(\beta\)-VAE with \(\beta\)=0.9). This result suggests that NE-VAE se produces smoother latent representation with potentially less saddle points and local minima.

\section{Discussion}

\subsection{Why vanilla VAE suffer a high re-encoding loss}

In our experiments, we observed that vanilla VAE will report a high re-encoding error, but reducing the weight on KL term will reduce re-encoding error even if we don't explicitly constraint the loss on it, as shown in fig-\ref{se-training}. We hypothesize that the high re-encoding error isn't because vanilla VAE is incapable of encoding samples that's close in \(\mathcal{X}\) to be close in the latent space. Instead, VAE may only maintain this property in the active dimensions, while the collapsed dimensions will report a large re-encoding error, resulting in a overall large re-encoding error. This behavior also may explain why constraining for re-encoding error can potentially force the model to make use of all of the latent dimensions.

This result also suggests that re-encoding error maybe an valuable metric for measuring the phenomenon of posterior collapse.

\subsection{Importance of additional constraints for optimal latent code}

From our experiments, it's quite obvious that VAE variants that gives similar ELBO and KL scores, or even similar measured mutual information included, may still behave quite differently in the latent space. This may be a concern for specific downstream tasks, for which vanilla VAEs will very likely give latent space of sub-optimal performance. To remedy this problem, researchers may benefit from using different forms of latent representation constraints, or link the training of the variational autoencoder and constraint the organization of latent space directly on the performance of downstream tasks. At this preliminary stage, it's not clear yet if there exists universally optimal latent organization that can be wildly applied to all applications.

\subsection{Roles of data configuration in posterior collapse}

We observe different latent activity distribution on different data set with same model, which suggests data configuration may play a role in the phenomenon of posterior collapse. Since MNIST and Omniglot data set used are similar in image quality and sample quantity, while having identical image size, we hypothesize that higher sample count per class in MNIST may be the contributing factor. If this can be proven the case, we can interpret the ability of NE-VAEs to retain more active dimensions as it has actively populated the data set \textit{post hoc} with samples generated by the decoder, thus enlarging per-class sample counts. Whether performance on few-shot or one-shot data set will benefit the most from NE-VAE requires further investigation.

\section{Conclusion}

In this paper we present the novel neighbor embedding VAE(NE-VAE) framework, which constraint the encoder to retain the topological feature of the input space in the latent space, and provided empirical evidence for its effectiveness of preventing posterior collapse. The ineffectiveness of simple numerical metric for measuring the performance and latent organization of VAEs suggest that downstream tasks that rely on the quality of latent representation may need to implement different constraints on the latent organization to reach optimal overall performance.


\bibliography{reference}
\bibliographystyle{icml2021}
\end{document}